# DeeP-Mod: Deep Dynamic Programming based Environment Modelling using Feature Extraction

Dr Chris Child and Lam Ngo

City St George's, University of London, London, EC1V 0HB, UK

**Abstract.** The DeeP-Mod framework builds an environment model using features from a Deep Dynamic Programming Network (DDPN), trained via a Deep Q-Network (DQN). While Deep Q-Learning is effective in decision-making, state information is lost in deeper DQN layers due to mixed state-action representations. We address this by using Dynamic Programming (DP) to train a DDPN, where value iteration ensures the output represents state values, not state-action pairs. Extracting features from the DDPN preserves state information, enabling task and action set independence. We show that a reduced DDPN can be trained using features extracted from the original DDPN trained on an identical problem. This reduced DDPN achieves faster convergence under noise and outperforms the original DDPN. Finally, we introduce the DeeP-Mod framework, which creates an environment model using the evolution of features extracted from a DDPN in response to actions. A second DDPN, which learns directly from this feature model rather than raw states, can learn an effective feature-value representation and thus optimal policy. A key advantage of DeeP-Mod is that an externally defined environment model is not needed at any stage, making DDPN applicable to a wide range of environments.

**Keywords:** Deep Dynamic Programming Network, Neural Network, Dynamic Programming, Feature Extraction, Feature-based Model, Environmental Modelling, Deep Q-Learning.

## 1 Introduction

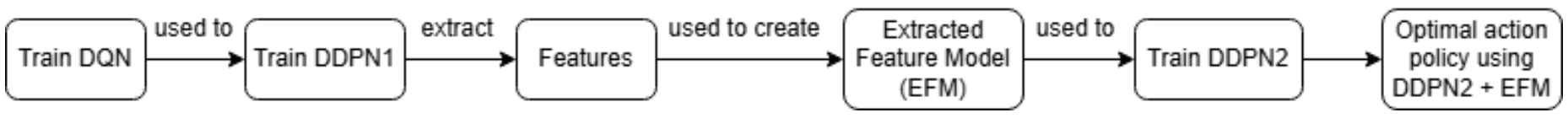

**Fig. 1.** DeeP-Mod is an environment modelling framework, where states are converted to features and an extracted feature model (EFM) created, mapping [features + action] to [features'].

Reinforcement Learning (RL) uses rewards and penalties to train agents through trial and error, mapping state-action pairs to expected future rewards [1]. This guides agents to maximize cumulative rewards. Deep Learning (DL) is a category of Machine Learning algorithms that leverage artificial neural networks to automatically derive insights from raw data. A Deep Q-Network (DQN) [2] combines RL with DL, leveraging neural networks to extract meaningful features from raw data. Features extracted from hidden layers of neural networks can effectively be utilized to train a new DQN. These extracted features encapsulate critical representations that enhance the learning process, allowing the DQN to achieve improved performance [3].

Dynamic Programming (DP) is an alternative approach that optimizes control using state value functions updated with Bellman equation. A Deep Dynamic Programming Network (DDPN), unlike a DQN, builds a state value representation rather than a state-action value representation, improving its suitability for feature extraction. The output of this network is a state value for each state, which means that only state information is encoded in the hidden layers instead of combined state-action. One drawback of DP is that it can only solve decision-making problems where the environment's dynamics are known and can be modeled precisely. However, in many real-world environments a well-defined transition model is not known, making direct application of DP impossible. To address this, our work creates an environment model using features extracted from a DDPN trained on values estimated by a model-free DQN. These features are meaningful and compact representations derived from raw states. Instead of relying on explicit state definitions, state-action transitions are recorded in feature space, allowing a feature-based model to be trained using DP.

This work has two main goals. Firstly, we aim to train a DDPN using features extracted from a previously trained DDPN. The system will use features extracted from the third hidden layer of a DDPN and use these as inputs to a simplified DDPN that has fewer layers. Noisy inputs are also used to test the efficiency of the state information encoded in those features. These methods demonstrate that critical information is encoded and preserved in the hidden layers and thus can be used to enhance the learning performance of the agent. Secondly, we aim to create a framework for environment modelling using feature extraction in DDPN. This framework is called DeeP-Mod (**Dee**p Dynamic **P**rogramming based Environment **Mod**elling). A traditional model works by directly mapping the combination of state + action [$S^1$ + A] to the resulting state [$S^2$], recording every observed state-action pair and the next state that follows. In contrast, DeeP-Mod models features extracted + action [$F^1$ + A], to the resulting next state's features extracted [$F^2$] (Figure 1). This approach enables the application of DP in environments without predefined transition models.

The paper has four parts. The first part gives context, discussing research on feature extraction, reinforcement learning and dynamic programming. Detailed methods and the problem statement are discussed in part two. Part three presents the results, which includes tables showing the accuracy of the derived state values, and graphs to show agent learning performance. Lastly, the conclusion provides analysis, possible extensions, and overall conclusions about the research.

## 2 Context

A Deep Learning Network includes multiple hidden layers between the input and output layer [5]. The network learns data representations across multiple layers of abstraction by utilizing these hidden layers [6]. At each hidden layer, there is an activation function, such as hyperbolic tangent (*tanh*), or rectifier (*ReLU*). These activation functions are applied to the weighted sum of the units from the previous layer to get a new representation of data [5]. "Inceptionism" [7] and "DeepDream" [8] demonstrated that when a Deep Neural Network is trained with a large dataset of related images and its parameters are adjusted, each layer of the network learns increasingly complex features of the image. Early layers capture basic features like edges and corners, while intermediate layers detect simple shapes such as leaves or doors. In the final layers, these simpler elements are combined to recognize more complex structures, e.g., buildings or trees, culminating in the output layer, which classifies or interprets the image as a whole. Previous work showed that an agent trained using features extracted from the hidden layers of the Deep Q-Network performs faster by an average factor of 4.58, implying that these layers encode critical environmental information which can be leveraged to optimize action selection [3].

Feature extraction can be used to enhance the efficiency of the learning processes of an agent [3]. In this research we use a value function algorithm (DP) instead of the Q-learning algorithm to train the agent. A value function predicts the cumulative and discounted feature rewards [5]. The state value represents the expected (E) cumulative reward the agent will achieve if it follows the policy from that state [5][1]. The state value under a policy $\pi$ is defined as [1]:

$$v_{\pi}(s) = E_{\pi}[\, R_t \mid s_t = s \,] \tag{1}$$

This satisfies the Bellman equation as [1]:

$$v_{\pi}(s) = \sum_a \pi(\, a \mid s\,) \sum_{s',r} p(\, s', r \mid s, a\,)[r + \gamma v_{\pi}(s')] \tag{2}$$

The optimal state value $v{*}(s) = max_{\pi}\, v_{\pi}(s)$ can be obtained using DP. DP can only be used to compute an optimal policy when a model of the environment is available, typically represented as a Markov Decision Process (MDP) [1]. MDP is a common framework for sequential decision-making and planning [9]. An MDP is defined by: States ($S$), Actions ($A$), Transition function ($P(s'|s,a)$), and Reward function ($R(s,a)$) [10]. State is the situation the agent is currently in. Actions are a set of choices available to the agent in a particular state. The transition function represents the probability of transitioning from one state to another given an action and a current state. The reward function determines the rewards the agent receives when taking an action, $a$, in state, $s$. The optimal policy can be obtained by finding the value function $v{*}(s)$ that satisfies the Bellman equation:

$$v^{*}(s) = \max_{a} \sum_{s',r} p(\, s', r \mid s, a\,)[r + v^{*}(s')] \tag{3}$$

The Bellman equation was introduced in the 1950s by Richard Bellman and is central to DP and its use to solve optimal control problems [4]. There are two common DP techniques: policy iteration and value iteration. This research uses the value iteration algorithm, which uses the Bellman equation as an update rule, iteratively refining approximations of the value function [1].

# 3 Method

## 3.1 Test Environment

The agent's task is to find the shortest path between two states in an environment while also avoiding obstacle states, called holes. The grid is 4x4 with one agent and three holes. This environment is inspired by the Frozen Lake problem from AI Gym [11]. As Figure 2 shows, the agent starts at position A and will need to find the optimal route to the final position, position P. The holes are in positions F, H, L and M. Possible actions are up, down, left, and right. In our experiment, the environment is deterministic.

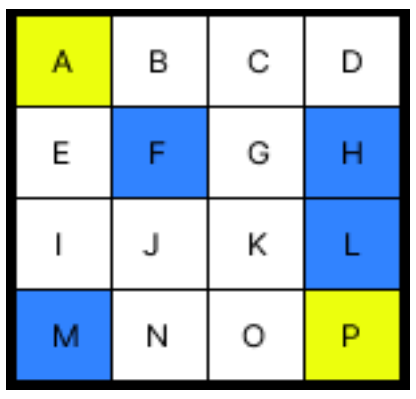

**Fig. 2.** The Frozen Lake environment (AI Gym [11]).

## 3.2 Dynamic Programming and Value Iteration

The value iteration algorithm is used to calculate the approximate value of each state. It iteratively refines the value of each state until it converges to the optimal value using Equation (3). In our implementation, we use a state-based reward formulation where rewards are attributed directly to current states rather than transitions. This means that $r$ is the immediate reward for being in state $s$. This approach is particularly well-suited for deterministic grid environments like our Frozen Lake variant, where the reward structure corresponds to specific locations: goal states provide positive rewards, hole states impose penalties, and regular states incur step costs. It also provides more intuitive reward attribution where values directly reflect the desirability of being in specific states.

Once the value function converges, the optimal policy ($\pi*$) can be determined by applying the following rule, which identifies the best action to take in each state:

$$\pi^*(s) = \arg\max_{a} \sum_{s',r} p(s', r \mid s, a)[r + \gamma V(s')] \quad (4)$$

## 3.3 Deep Dynamic Programming Network

The Deep Dynamic Programming Network (DDPN) integrates a Deep Neural Network with the Value Iteration algorithm to compute optimal policies for an agent. It takes as input a one-hot encoded 1×16 vector representing each state, and outputs a value for each state (Figure 3). This value function is then used to derive the optimal policy. The network is implemented using TensorFlow [12] and Keras.

The architecture consists of an input layer, five hidden layers, and an output layer. The first three hidden layers each contain 32 neurons and use the hyperbolic tangent *(tanh)* activation function. The last two hidden layers also have 32 neurons but use the rectified linear unit *(ReLU)* activation function. The network is configured with the following parameters: a discount factor $\gamma$=0.9, the Adam optimizer [13], and the Mean Squared Error (MSE) loss function.

During training, the agent undergoes 200 episodes, which is sufficient for the value function to converge. Initially, each state's value is set to zero. Values are recursively updated using the Bellman equation with our state-based reward formulation as detailed in Section 3.2. Every two iterations, the optimal policy ($\pi^*$) is executed to assess training progress. The rewards are: +10 for reaching the goal, -1 for each step taken, and -10 for falling into a hole. After training is complete, the optimal policy ($\pi^*$) is tested again to evaluate performance.

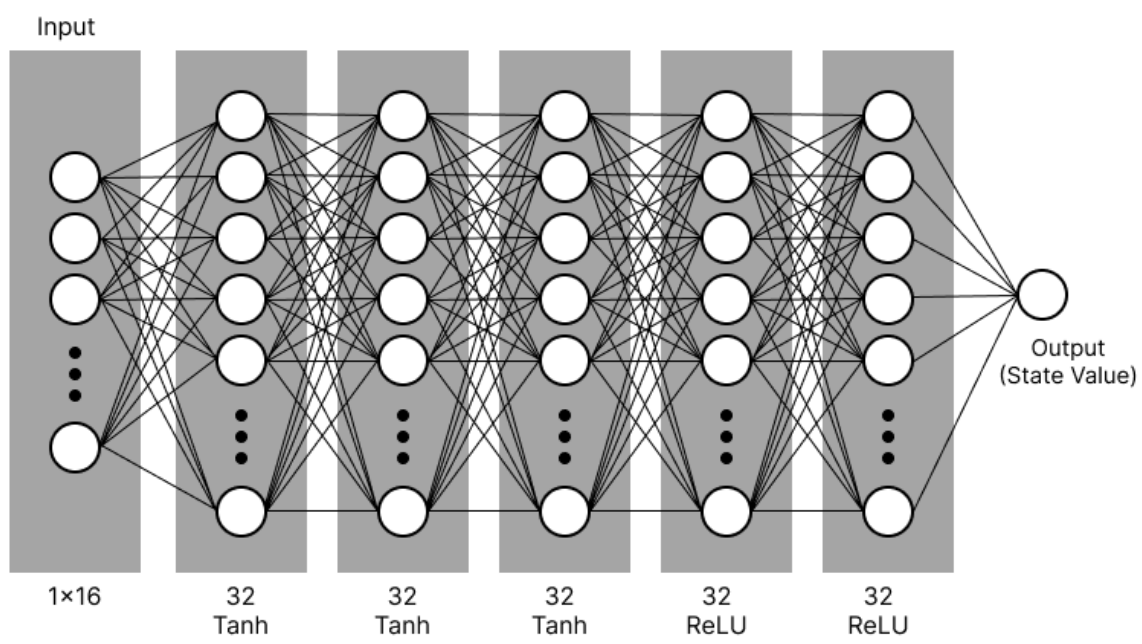

**Fig. 3.** The original DDPN with 16 inputs, 5 hidden layers and 1 output value.

### 3.4 Noisy Input

Real-world environments contain irrelevant information, imperfect information, or noise. Adding noisy inputs can simulate these issues so that we can test how the agent performs in such situations. Noise was added as 20 extra features (random 0s and 1s) to the original 1×16 encoded state vector, resulting in a 1×36 input. The noisy DDPN structure also has 5 hidden layers that have the same number of neurons and activation functions as the original network (Figure 4 [left]).

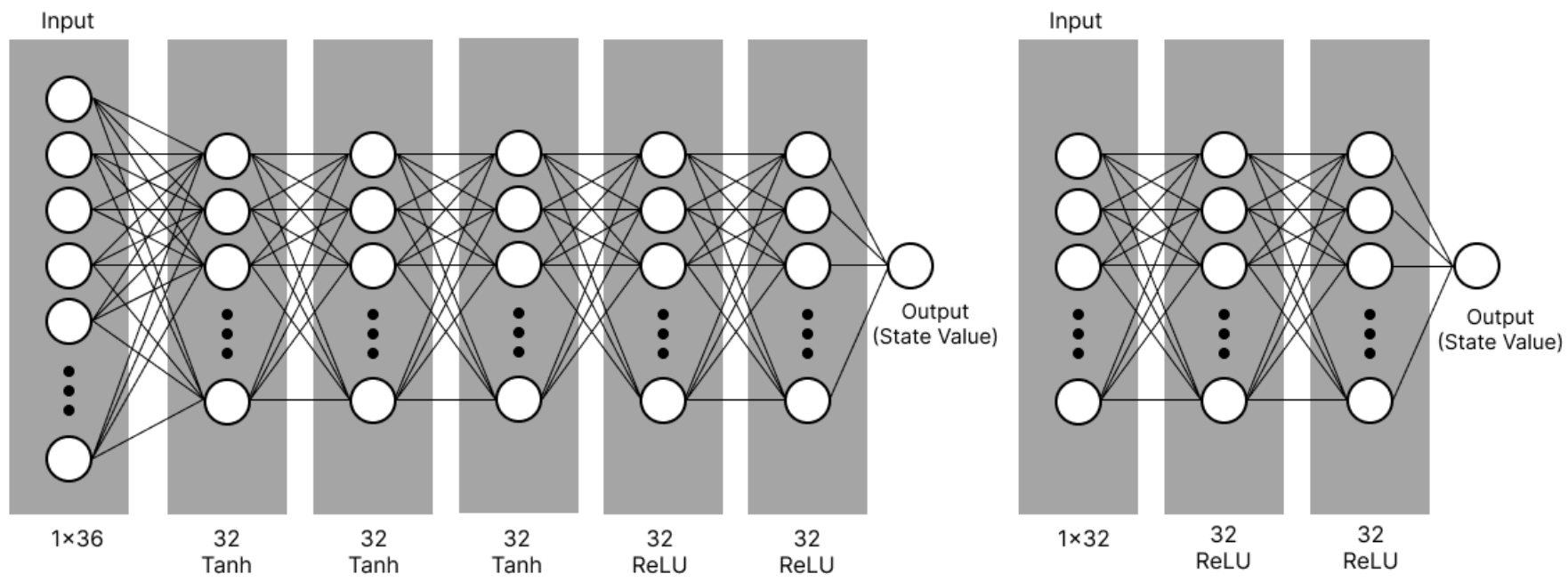

**Fig. 4.** The DDPN with 32 input neurons (right) will be trained using features extracted from the third hidden layer of the DDPN with 36 input neurons (left) as inputs.

### 3.5 Feature Extraction from Third Hidden Layer of Deep Dynamic Network

Features are extracted from the neurons of the third hidden layer of the DDPN for both no-noise (16×32×32×32×32×32×1) and noisy (36×32×32×32×32×32×1) cases. This layer is defined with the *tanh* activation function and the output from the feature extraction layer is in the range of -1 to 1. If the output of a neuron is greater than 0, it is represented as 1, and if it is less than 0, it is represented as -1. This encoding captures the state information when extracting features from the network. The activation of this layer is used as an input for a new DDPN. As the third hidden layer has 32 neurons, the input layer of the new DDPN takes encoded 1×32 vectors as input. The new DDPN has 2 hidden layers; each layer also has 32 neurons with *ReLU* activations, and its output is a state value. The architecture is the same for both no-noise and noise cases (Figure 4 [right]).

### 3.6 Feature-based Model Trained with DeeP-Mod

The DeeP-Mod framework consists of the following steps (**Fig. 1**): (i) apply Q-Learning to estimate the value of state-action pairs; (ii) train a DDPN (DDPN1) using state values extracted from the Q(s,a) table or DQN; (iii) extract features from a hidden layer of DDPN1; (iv) create an extracted feature model (EFM); (v) train a second DDPN (DDPN2) using the EFM; (vi) optimal action policy using EFM and DDPN2.

Step (i): Apply Q-learning to estimate the value of state-action pairs. We use both a Deep Q-Network (DQN) and a tabular Q-learning approach. In the tabular setting, Q-values are updated directly with learning rate α=0.9 and discount factor γ=0.9, using the following update rule:

$$Q(s,a) = Q(s,a) + \alpha[r + \gamma a' \max Q(s',a') - Q(s,a)] \quad (5)$$

For DQN, value estimation is performed by training a neural network, using γ=0.9, the Adam optimizer (with learning rate set via the optimizer, α=10-4) [13] and the MSE loss function.

Step (ii): Train a DDPN (DDPN1) using state values extracted from the Q(s,a) table or DQN. The state value is extracted by selecting the value of a state-action pair that leads to that state. In our experiments, we use the same architecture as the DDPN in Section 3.4 (left).

Step (iii): Extract features from the third hidden layer of the DDPN1, following the method outlined in Section 3.5. Each state s is mapped to its corresponding feature vector f, and the mappings *(s→f)* are stored in a dictionary for efficient lookup during model construction in next step.

Step (iv): Construct an Extracted-Feature-based transition Model (EFM) by interacting with the environment through a combination of random exploration and policy-directed actions. For each state *s*, its corresponding feature representation *f* is obtained from the state-to-feature mapping, previously generated by extracting features from the third hidden layer of DDPN1. From each state, the agent selects an action a ∈ A (|A|=4 in our setup) using an epsilon-greedy strategy. The process begins with a high epsilon value (ε=0.9) and an epsilon decay rate of 0.99 per episode. This allows the agent to balance between taking random action to explore the environment fully and policy-directed action to explore areas that might be hard to reach.

Upon taking an action *a*, the environment returns the next state *s'*, and associated reward *r*. The feature representation *f'* of the resulting state *s'* is retrieved using the previously constructed state-to-feature mapping. Each transition *(f,a)→(f',r)* is stored in the feature-based transition table, forming the EFM. The process is repeated across multiple episodes until all state–action pairs have been covered at least once. The resulting EFM captures how feature representations evolve under different actions. The model is represented as a lookup table in this work but could equivalently be captured by a Deep Neural Network (DNN) representation. Note: in stochastic environments there will be multiple *f'* and a sample-based model can be constructed.

Step (v): Train a second DDPN (DDPN2) using the EFM and features as inputs. Given a feature representation *f*, the agent selects action *a*, looks up the next feature representation *f'*, with reward *r*, from the EFM, and updates its value function for *f'* using DP. In our experiments, the DDPN2 input is a 1×32 vector and the network has two hidden layers, each with 32 neurons and *ReLU* activations.

Step (vi): Optimal action policy using EFM and DDPN2. In our experiments an optimal policy is calculated as in Section 3.3 to evaluate the performance in the training and testing phase.

# 4 Results

## 4.1 State Values for Frozen Lake (no Noise Inputs)

Table 1 shows the value of each state after training the agent for 200 episodes with the original input of size 1×16. Light grey cells indicate hole states, and dark grey cells indicate the goal state.

The state value comparison indicates that both the DDPN and the reduced DDPN produce values closely aligned with tabular value iteration.

**Table 1.** Value comparison for each state for DDP training with tabular value iteration, DDPN (no noise inputs), reduced DDPN (no noise inputs).

| State | Value Iteration | DDPN | Reduced DDPN |
|---|---|---|---|
| A | 5.31 | 5.43 | 5.31 |
| B | 5.91 | 5.74 | 5.42 |
| C | 6.56 | 6.48 | 6.11 |
| D | 5.91 | 5.72 | 5.43 |
| E | 5.91 | 5.73 | 5.47 |
| F | -3.44 | -3.46 | -3.92 |
| G | 7.29 | 7.20 | 6.99 |
| H | -3.44 | -3.50 | -3.80 |
| I | 6.561 | 6.53 | 6.24 |
| J | 7.29 | 7.22 | 6.93 |
| K | 8.01 | 7.88 | 7.74 |
| L | -1.01 | -1.13 | -1.41 |
| M | -2.71 | -2.77 | -3.10 |
| N | 8.01 | 7.92 | 7.72 |
| O | 8.99 | 8.83 | 8.60 |
| P | 9.99 | 9.85 | 9.62 |

### 4.2 State Values for Frozen Lake with Noisy Inputs

Table 2 shows the state values after training the agent for 200 episodes with the noisy input vector of size 1×36. Light grey cells indicate hole states, and dark grey cells indicate the goal state.

The value comparison across different states indicates that noisy DDPN (Section 3.4, left) assigns higher values than the tabular value iteration across all states. On the other hand, the reduced DDPN (Section 3.4, right) produces values that are almost identical to the tabular methods. DeeP-Mod (feature-based) DDPN also outputs values that are close to the tabular value iteration in most cases.

**Table 2.** Value comparison for each state after training DDP with tabular value iteration, DDPN (noisy inputs), reduced DDPN (noisy inputs), and DeeP-Mod DDPN (noisy inputs).

| State | Value Iteration | DDPN | Reduced DDPN | DeeP-Mod DDPN |
|---|---|---|---|---|
| A | 5.31 | 5.43 | 5.31 | 5.60 |
| B | 5.91 | 6.26 | 5.90 | 6.65 |
| C | 6.56 | 6.91 | 6.56 | 6.87 |
| D | 5.91 | 5.97 | 5.90 | 6.59 |
| E | 5.91 | 6.37 | 5.90 | 6.07 |
| F | -3.44 | -2.52 | -3.44 | -3.49 |
| G | 7.29 | 7.66 | 7.29 | 7.54 |
| H | -3.44 | -3.12 | -3.44 | -3.50 |
| I | 6.561 | 7.31 | 6.561 | 6.96 |
| J | 7.29 | 7.77 | 7.29 | 7.87 |
| K | 8.01 | 8.54 | 8.099 | 8.64 |
| L | -1.01 | -0.32 | -1.00 | -0.76 |
| M | -2.71 | -2.34 | -2.709 | -2.45 |
| N | 8.01 | 8.93 | 8.09 | 8.48 |
| O | 8.99 | 9.81 | 8.99 | 9.39 |
| P | 9.99 | 10.49 | 9.99 | 10.3 |

### 4.3 Performance for Noisy DDPN

Training the DDPN 36×32×32×32×32×32×1 took 1400.46 seconds, while using features from the third hidden layer in the reduced DDPN 32×32×32×1 improved training time to 677.64 seconds. The test was run on a Lenovo Legion Pro 5 laptop (Windows 11 Home 64-bit, Intel Core i9-14900HX CPU @ 2.2 GHz, 32.0GB RAM, Ge-Force RTX 4070).

Figure 5 compares the rewards of DDPN 36×32×32×32×32×32×1 (trained with noise) and DDPN 32×32×32×1 (using third-layer feature extraction). During training, DDPN 36×32×32×32×32×32×1 takes longer to stabilize (~75 iterations) with more fluctuation, while DDPN 32×32×32×1 improves rapidly, stabilizing around iteration 25. In testing, DDPN 32×32×32×1 maintains a steady 4.0 reward, while DDPN 36×32×32×32×1 shows occasional drops, indicating less consistency due to noise.

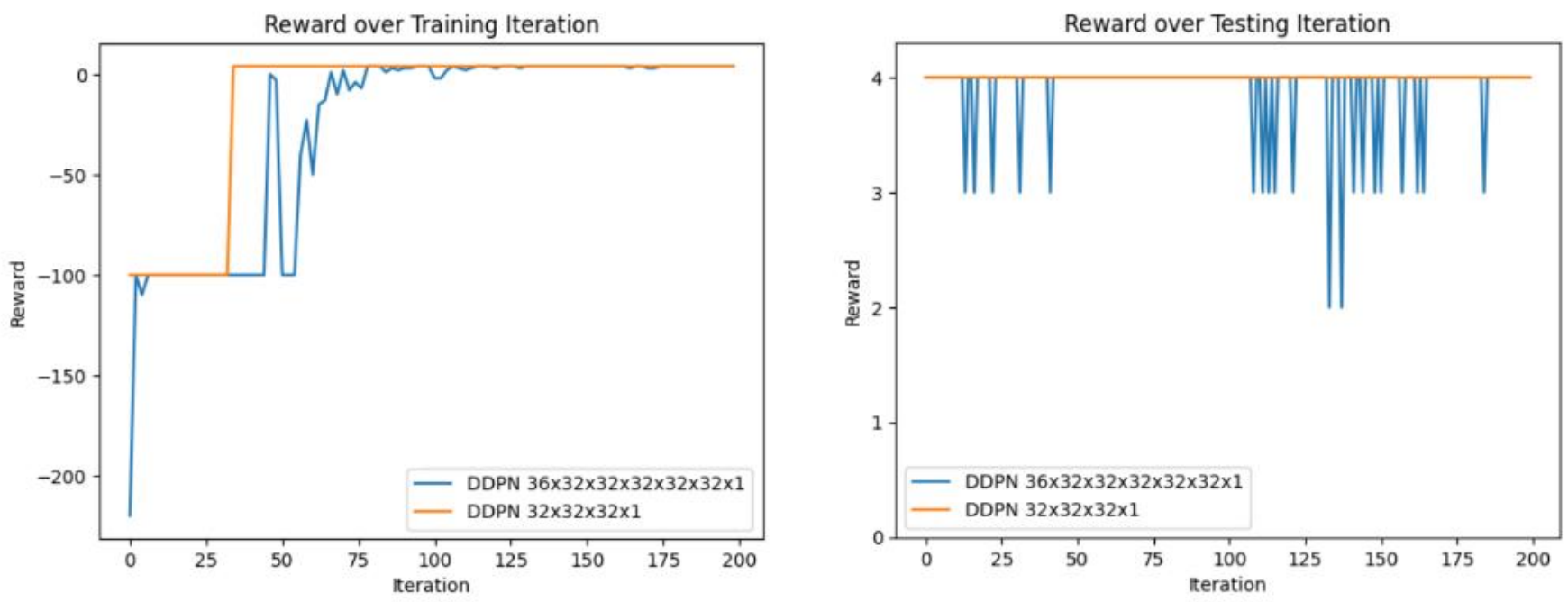

**Fig. 5.** Reward between DDPN 36x32x32x32x32x32x1 and DDPN 32x32x32x1 over training iteration and testing iteration (20 noise inputs, $\gamma = 0.9$)

Figure 6 compares the rewards during training and testing for DDPN 16×32×32×32×32×32×1 (original with no noise inputs) and DDPN 32×32×32×1. In training, both models start with low rewards (-100) as they initially favor certain states without progressing. After a steep reward increase (~25 iterations), they stabilize as state values improve, achieving the same reward of 4.0. In testing, both maintain a constant reward of 4.0 throughout 200 iterations, indicating full training and consistent policy selection based on state values.

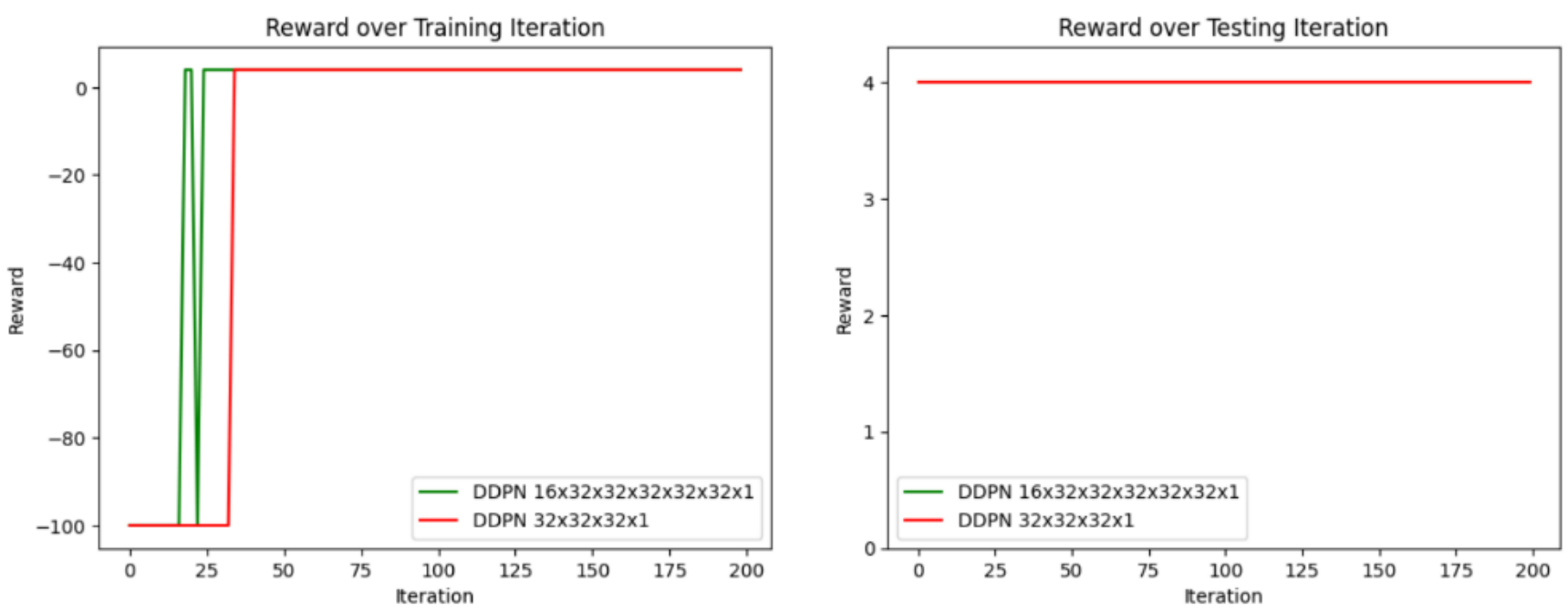

**Fig. 6.** Reward between DDPN 16x32x32x32x32x32x1 and DDPN 32x32x32x1 over training iteration and testing iteration (no noise inputs, $\gamma = 0.9$).

### 4.4 Performance for Feature-based Model: DeeP-Mod

Figure 7 compares rewards during training and testing for two models: the DDPN 36×32×32×32×32×32×1 trained with noisy inputs and Q-values, and DeeP-Mod (feature-based DDPN) 32×32×32×1 trained with a feature transition model. During training, the DDPN 36×32×32×32×32×32×1 shows initial fluctuations and stabilizes around iteration 50. In contrast, the feature based DDPN 32×32×32×1 rapidly in-creases rewards by iteration 25 and stabilizes at 4.0. In testing, the feature based DDPN maintains a stable reward of 4.0, while the DDPN 36×32×32×32×32×32×1 experiences occasional drops, indicating sensitivity to noise.

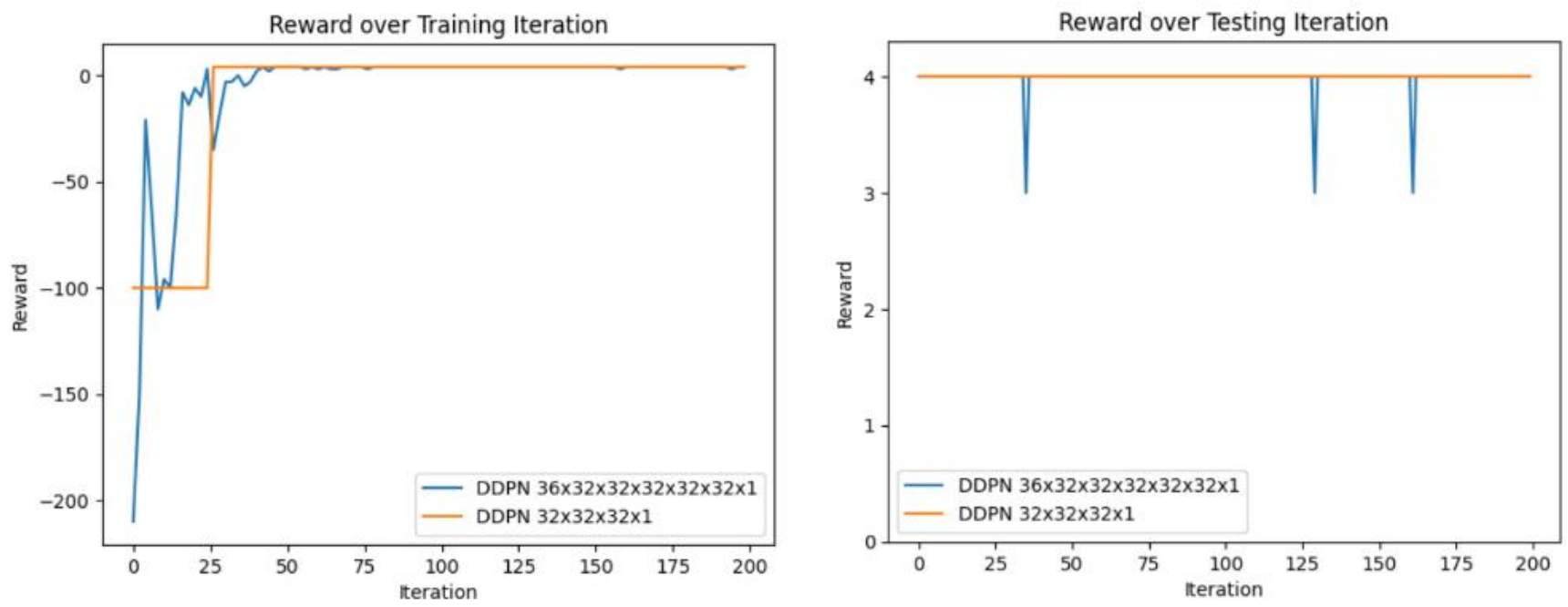

**Fig. 7.** Reward between DDPN 36x32x32x32x32x32x1 and DeeP-Mod (feature-based DDPN) 32x32x32x1 over training iteration and testing iteration (20 noise inputs, $\gamma$ = 0.9).

## 5 Conclusion

This work demonstrates that feature extraction from a DDPN can enhance agent learning by extracting key features from the input while filtering out irrelevant in-formation. The DDPN generates state values rather than state-action values, eliminating action information from the network, making it possible to extract state-feature information. This will allow features to be used to solve novel problems based on the same states. When training the network with noisy inputs, the reduced DDPN filters out the noise and focuses on important state information. When features are extracted from the third hidden layer and used to train a new model, this new model not only produces correct values for each state but also exhibits similar performance to the original model trained in a noise-free environment. This similarity demonstrates that the extracted features successfully filter out the noisy in-puts, enabling the new model to achieve a learning speed comparable to that of the original no-noise environment, whereas learning in a noisy environment without feature extraction takes significantly longer. The new model trained with the third layer's extracted features achieves a faster learning rate and better overall performance compared to the original noisy network. This reduced network is not only able to disregard noise effectively but also demonstrates similar efficiency to the noise-free environment. Together, these findings validate that feature extraction enables the model to effectively disregard noise, improve learning efficiency, and achieve strong performance, making it a robust and efficient approach for training in noisy environments.

The work also introduced the DeeP-Mod framework, which creates a predictive environment based on the evolution of extracted features in response to actions (EFM). This approach can be used to extend the use of DP beyond environments with predefined state transitions. The framework combines DQN and DP for feature extraction.

Our initial work shows that the framework performs well on a small deterministic problem, with learning in feature space improving efficiency and providing a framework for future environments which do not have explicit state transition functions.

Future improvements could include: modifying the environment to swap the goal and hole states (demonstrating that feature extraction captures environment information independent of reward); replacing the feature-based transition table model with a neural network; experimenting with different model parameters, optimizers, activation functions and loss functions to further enhance performance and convergence.

The principles of the DeeP-Mod framework have been shown to work on a small-scale environment. The next steps are: testing the framework on simple stochastic environments; complex deterministic environments; complex stochastic environments. The iterative process of extracted feature environment modelling (EFM) using a large run of the original environment should be possible in such cases, leading to a wide range of new applications for Dynamic Programming.